%% file: arxiv.tex
\pdfoutput=1
\documentclass{article}
\usepackage{times,authblk}
\usepackage{fullpage}

\usepackage{geometry}
\usepackage[colorlinks,bookmarks=false]{hyperref}
\usepackage{rotating, color,subfigure,algorithm,algorithmic}

\usepackage{multirow}
\usepackage{graphicx}

\usepackage{enumerate}
\usepackage{amsthm,amsmath,amssymb}

\usepackage{mkolar_definitions}

\usepackage{ifthen}

\newcommand{\version}{arxiv}
\ifthenelse{\equal{\version}{arxiv}}{
\newcommand{\figsizeb}{0.2}
}{
\newcommand{\figsizeb}{0.24}
}
\begin{document}
\title{Subspace Learning from Extremely Compressed Measurements}

\author[1]{
Akshay Krishnamurthy
\thanks{akshaykr@cs.cmu.edu}}
\author[2]{
Martin Azizyan
\thanks{mazizyan@cs.cmu.edu}}
\author[2]{
Aarti Singh
\thanks{aarti@cs.cmu.edu}}

\affil[1]{Computer Science Department\\
Carnegie Mellon University}
\affil[2]{Machine Learning Department\\
Carnegie Mellon University}

\maketitle

\input{abstract.tex}
\input{intro.tex}
\input{results.tex}
\input{experiments.tex}
\input{conclusion.tex}

\section*{Acknowledgements}
This research is supported in part by NSF awards IIS-1116458, IIS-1247658 and CAREER IIS-1252412 and an AFOSR YIP award.
AK is supported in part by a NSF Graduate Research Fellowship.

\bibliography{mc}
\bibliographystyle{plain}

\vfill
\newpage 
\appendix
\input{appendix.tex}
\end{document}

%% file: abstract.tex
\begin{abstract}
We consider learning the principal subspace of a large set of vectors from an extremely small number of compressive measurements of each vector. 
Our theoretical results show that even a constant number of measurements per column suffices to approximate the principal subspace to arbitrary precision, provided that the number of vectors is large. 
This result is achieved by a simple algorithm that computes the eigenvectors of an estimate of the covariance matrix.
The main insight is to exploit an averaging effect that arises from applying a different random projection to each vector. 
We provide a number of simulations confirming our theoretical results. 
\end{abstract}

%% file: intro.tex
\section{Introduction}
\label{sec:intro}

Low rank approximation is a fundamental preprocessing task in a wide variety of machine learning and signal processing applications.
Given a set of vectors $x_1, \ldots, x_n \in \RR^d$, low rank approximation, popularly referred to as principal component analysis, aims to find a low dimensional subspace of $\RR^d$ that captures a large amount of the energy present in the vectors $\{x_t\}_{t=1}^n$. 
It is well known that if we concatenate the vectors as columns of a $d \times n$ matrix $X$, then the span of the left singular vectors corresponding to the $k$ largest singular values of $X$ forms the best $k$-dimensional subspace in many senses. 
For example, it minimizes the squared Euclidean reconstruction error, $|| X - \Pi X||_F^2$, over all $k$-dimensional projections $\Pi$.


There are many situations where obtaining this best subspace is computationally challenging, motivating several different theoretical and empirical studies. 
A number of clever approaches, ranging from sparsification~\cite{achlioptas2007fast} and column sampling~\cite{boutsidis2011near}, to sketching and streaming techniques~\cite{liberty2013simple,drineas2003pass} have been used to reduce the computational burden associated with computing the principal subspace when $n$ and $d$ are large.

An equally pressing concern is the cost of data acquisition, motivating lines of work on low rank approximation from missing data and compressive measurements. 
The matrix completion literature focuses on finding approximations from randomly or adaptively sampled entries of the data matrix~\cite{recht2011simpler,negahban2012restricted,krishnamurthy2013low,gonen2014sample}.
With compressive measurements, only a few linear combinations of each column of $X$ are acquired, which mathematically translates to observing $Y = RX$ where $R$ is a $m \times d$ random matrix with $m \ll d$. 
Usually, the columns of $Y$ are projected back into $d$-dimensions and the principal subspace of these vectors is used as an estimator~\cite{sarlos2006improved,halko2011finding,liberty2007randomized}.
In this paper, we deviate from this strategy, and propose an algorithm that uses a different sensing operation per column to approximate the principal subspace of the vectors $\{x_t\}_{t=1}^n$.

This difference, on one hand, adds some complexity to the algorithm, as a new random projection must be applied to each column.
On the other hand, resampling the projection leads to an averaging phenomenon that allows us to undersample each column significantly more than existing approaches.
In particular, we show that our algorithm only needs two compressive measurements per column provided there are enough columns, while existing approaches require $O(k/\epsilon)$ measurements to achieve an $\epsilon$-approximation with target rank $k$~\cite{halko2011finding}. 

Before proceeding, let us briefly mention two motivating applications. 
In time-series analysis, typically we have extremely long data sequences and would like to extract some signal from the sequence to assist in some inferential task.
In many of these problems, observing any time-point of the data sequence is expensive; it may require performing some scientific experiment.
Our results show that one can significantly undersample each time point of the series, which can tremendously reduce measurement overhead, while still extracting the principal subspace from the data sequence. 

A related application is in signal processing over a distributed sensor network with high measurement and communication cost.
Suppose each of the $n$ sensors records a $d$-dimensional vector of measurements, and our goal is to make inferences about the $d \times n$ matrix of measurements. 
Since measurement costs are high, one might be interested in compressive sampling at each sensor, but if the compression operator is shared across the sensors, an expensive synchronization step must be performed before data acquisition.
Our method avoids the need for synchronization, since each sensor uses its own compression operator, while guaranteeing a good approximation to the matrix of measurements.


%% file: results.tex
\section{Results}
\label{sec:results}

We first set up some notation used throughout the manuscript.
We are interested in recovering the principal $k$-dimensional subspace $\Pi$ of the set of vectors $x_1, \ldots, x_n \in \RR^d$ which we will concatenate into a matrix $X \in \RR^{d \times n}$. 
We assume that each vector $x_t$ has bounded norm, that is $\max_{t \in [n]} ||x_t||_2^2 \le \mu$\footnote{We use $[n]$ to denote $\{1, \ldots n\}$.}.
Let $\Sigma = \frac{1}{n} \sum_{t=1}^n x_t x_t^T$ be the covariance matrix and notice that the optimal $k$-dimensional subspace $\Pi$, in terms of minimizing squared Euclidean reconstruction error, is spanned by the top-$k$ eigenvectors of $\Sigma$. 
A key parameter that governs the performance of our algorithm is the \textbf{eigengap} $\gamma_k$, defined as the difference between the $k$th largest eigenvalue and the $k+1$st largest eigenvalue of the covariance matrix $\Sigma$. 

We measure the error of an estimate $\hat{\Pi}$ for the principal subspace in terms of the spectral norm $||\hat{\Pi} - \Pi||_2$, where $||M||_2$ is the largest singular value of $M$. 
It is not too hard to show that when $\hat{\Pi}$ and $\Pi$ are projection matrices, $||\hat{\Pi} - \Pi||_2$ corresponds to the sine of the largest principal angle between the two subspaces, and therefore it is an appropriate measure of error in our setting. 

We are interested in a compressive sensing framework, in which for some $m \ge 1$, we observe each vector $x_t$ through $2m$ compressive measurements $y_{it} = a_{it}^Tx_t, i \in [2m]$ for vectors $a_i$ drawn independently from the $d$-dimensional unit sphere. 
This is equivalent to choosing a $2m$-dimensional projection $\Phi_t \in \RR^{d \times d}$ uniformly at random and observing $y_t = \Phi_t x_t$. 
In our algorithm, we will write this as two $m$-dimensional random projections $\Phi_t$ and $\Psi_t$ per column with observations $y_t = \Phi_t x_t$ and $z_t = \Psi_t x_t$. 
To connect the two representations, let $A_t$ denote the $d \times m$ matrix whose columns are the vectors $a_{it}, i \in [m]$ and define $\Phi_t = P_{A_t}$, the the projection onto the span of the columns $A_t$. 
Define $B_t$ similarly to $A_t$, but with the second $m$ vectors and let $\Psi_t = P_{B_t}$.

\begin{algorithm}[t]
\begin{algorithmic}
\STATE \textbf{Input:} Compression parameter $m$, target rank $k$.
\STATE $\hat{\Sigma} = 0 \in \RR^{d \times d}$
\FOR{$x_t \in \RR^d$ in the data stream}
\STATE Let $\Phi_t, \Psi_t$ be $m$-dimensional random projections.
\STATE Acquire $y_t = \Phi_t x_t$ and $z_t = \Psi_t x_t$.
\STATE Update $\hat{\Sigma} = \hat{\Sigma} + \frac{1}{2}(y_tz_t^T + z_ty_t^T)$
\ENDFOR
\STATE \textbf{Output:} $\hat{\Pi}= \textrm{span}(u_1, \ldots, u_k)$ the top $k$ eigenvectors of $\hat{\Sigma}$.
\end{algorithmic}
\caption{Compressive Subspace Learning}
\label{fig:algorithm}
\end{algorithm}

Our algorithm, which we call Compressive Subspace Learning (CSL), is conceptually quite simple (See Algorithm~\ref{fig:algorithm}). 
For each vector $x_t$, we observe $y_t = \Phi_t x_t, z_t= \Psi_t x_t$ and form an estimate for $C_t = x_t x_t^T$ with:
\begin{align}
\hat{C}_t = \frac{1}{2} (y_tz_t^T + z_ty_t^T)
\label{eq:chat_t}
\end{align}
We estimate the covariance $\Sigma$ with $\hat{\Sigma} = \frac{1}{n} \sum_{t=1}^n \hat{C}_t$ and use the span of top $k$ eigenvectors of $\hat{\Sigma}$ to estimate for the principal subspace $\Pi$. 
Let $\hat{\Pi}$ denote the output of our algorithm, namely the span of the top $k$ eigenvectors of $\hat{\Sigma}$.
Note that $\hat{\Sigma}$ needs to be appropriately rescaled before it provides a reasonable estimate of $\Sigma$, but since we are only interested in the principal subspace, this normalization is only necessary for analysis.

In addition to the statistical guarantee in Theorem~\ref{thm:main} below, we also mention some practical considerations. 
The algorithm can be implemented in the streaming model with each column vector $x_t$ being streamed through memory. 
In this model, the algorithm requires $O(d^2)$ time to process each vector and $O(d^2)$ space in total, to store $\hat{\Sigma}$.

The distributed sensor network model we described earlier is quite appealing for CSL.
Recall that we had a network of $n$ sensors, each observing a $d$-dimensional vector $x_t$, with high measurement and communication cost. 
A na\"{i}ve procedure that observes and transmits the vectors in full has $O(nd)$ communication overhead, but also $O(nd)$ measurement cost, which can be prohibitively expensive.
The traditional compressive sampling approach of using a single random projection needs only $O(mn)$ measurements but synchronizing the projection before data acquisition requires $O(nmd)$ communication cost. 
This is non-negligible overhead as one typically requires $m \asymp k/\epsilon$ to achieve error $\epsilon$ with target rank $k$~\cite{halko2011finding}.

Using CSL, each sensor can generate its own random projection, take $2m$ measurements of its signal $x_t$, and send only two $d$-dimensional vectors (namely $y_t$ and $z_t$) over the network. 
Thus, CSL makes $O(mn)$ measurements while suffering only $O(nd)$ communication overhead, achieving the best properties of both other approaches.
Moreover, provided $n$ is large, CSL can succeed even with $m$ constant, which is significantly smaller than other methods.

Our main result is the following statistical guarantee on the performance of the algorithm:
\begin{theorem}
With probability $\ge 1-\delta$:
\begin{align}
||\hat{\Pi} - \Pi||_2 \le \frac{1}{\gamma_k}\left(\sqrt{\frac{14\mu^2d}{nm}\log(d/\delta)} + \frac{2}{3}\frac{\mu d^2}{m^2n}\log(d/\delta)\right)
\label{eq:err_bound}
\end{align}
So that one can achieve spectral norm error $\le \epsilon$ provided that:
\begin{align}
n \ge \max\left\{ \frac{56 \mu^2 d \log(d/\delta)}{m \gamma_k^2\epsilon^2}, \frac{4}{3}\frac{\mu d^2}{\gamma_k \epsilon m^2}\log(d/\delta)\right\}
\label{eq:sample_complexity}
\end{align}
\label{thm:main}
\end{theorem}
Before turning to the proof, some remarks are in order:

We are interested in the setting where $m$ is small compared to $d$, in which case the second term in Equation~\ref{eq:err_bound} is active.
Thus, in interpreting the theorem, one should focus on the second terms in both bounds to see the dependence on $d$ and $m$. 
On the other hand, the relationship between $n$ and $\epsilon$ is dictated by the first terms. 

Note that since the theorem holds for any $m \ge 1$, even two compressive measurements per column suffice to approximate the principal subspace, provided that $n \gg d$, which is common in a number of signal processing and machine learning applications.
This is in sharp contrast with a number of other results on matrix approximation from compressive measurements, where the same projection is used on each column, in which case such a bound is not possible unless $m \ge k/\epsilon$~\cite{halko2011finding}. 
The justification for this is that using different random projections allows our algorithm to exploit averaging across the columns to capture the principal directions of the matrix. 
On the other hand, if the same projection is used, the measurements across columns become highly correlated and one does not experience a law-of-large-numbers phenomenon. 

\begin{figure}
\vspace{-0.25cm}
\begin{center}
\includegraphics[scale=0.3]{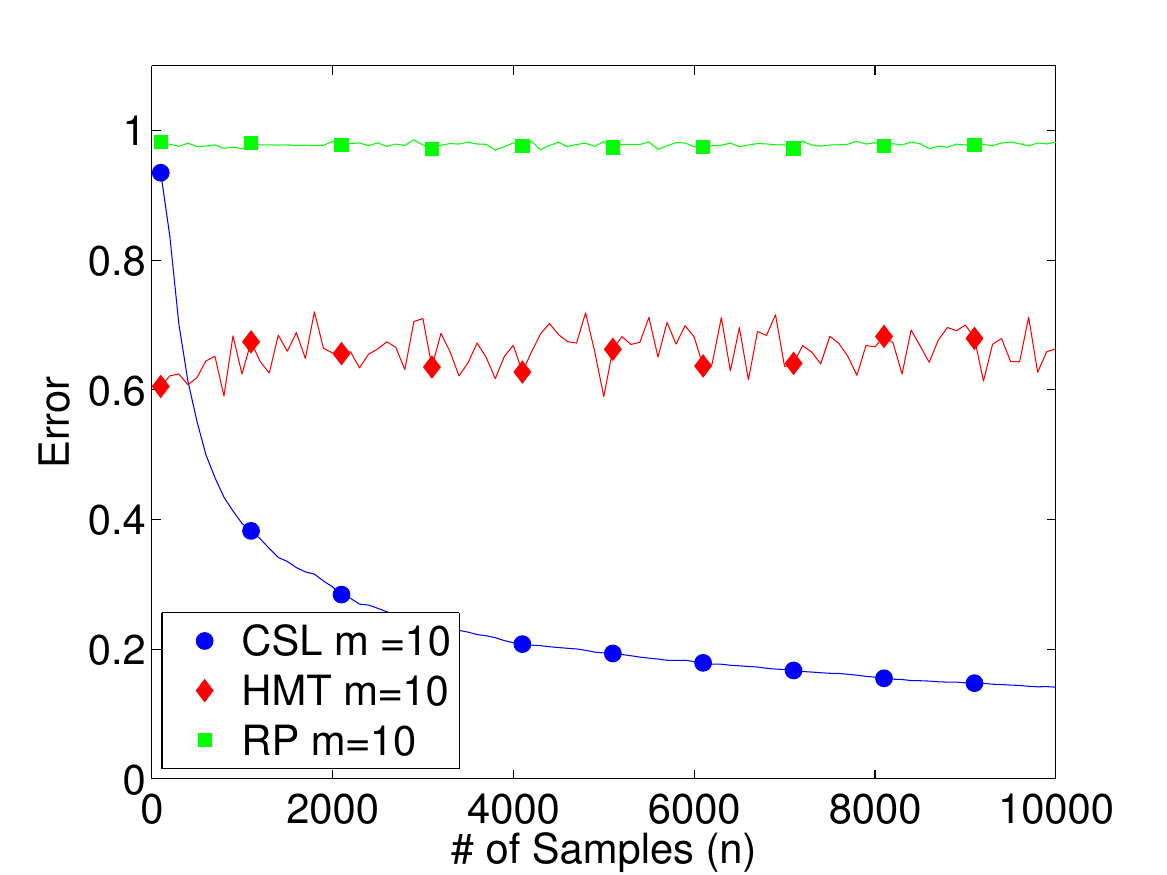}
\end{center}
\caption{Comparison of our algorithm with that of Halko \emph{et. al}~\cite{halko2011finding} (HMT) and the algorithm that uses a fixed random projection (RP) on a noisy low rank matrix $d=100, k=3$.}
\label{fig:hmt_compare}
\vspace{-0.5cm}
\end{figure}

To make this concrete, consider the setting where $X$ is a rank one matrix with identical columns.
If the same random projection is used across the columns, then there is no hope of approximating the principal direction with small $m$, irrespective of $n$. 
On the other hand, our bound says that by leveraging additional randomness in the data acquisition process, we can achieve an $\epsilon$-approximation provided that $n$ is large. 
In Figure~\ref{fig:hmt_compare} we demonstrate this phenomenon. 

Results of this flavor are usually stated in terms of the number of measurements per column that suffice to achieve error $\epsilon$.
Stated this way, our bound says that:
\[
m \asymp \max\left\{ \frac{d\mu^2\log(d/\delta)}{n\gamma_k^2\epsilon^2}, \frac{d}{\sqrt{n}} \sqrt{\frac{\mu}{\epsilon \gamma_k} \log(d/\delta)}\right\}
\]
is sufficient. 
In comparison, Halko \emph{et. al} only require $m \asymp k/\epsilon$ to achieve error $\epsilon$ with target rank $k$~\cite{halko2011finding} while the state-of-the-art results on matrix completion require observing $k \log(d/\delta)$ coordinates per column~\cite{recht2011simpler,negahban2012restricted}.
Both sets of results are better than ours when $k$ is small compared to either $d/n$ or $d/\sqrt{n}$, depending on which term is active. 
Note that the method of Halko \emph{et. al} does not compress the columns via a uniformly distributed projection, but rather projects onto a subspace computed by first compressing the rows of the matrix. 
While this approach can lead to better approximation, it is unfortunately not possible in many settings, including the distributed sensor network application mentioned above.


Our result is similar in spirit to a recent analysis by Gonen \emph{et. al} who study the subspace learning problem when only a small number of entries of each column are observed~\cite{gonen2014sample}.
They also show that one can approximate the principal subspace of a matrix $X$ using only a few observations per column.
Their analysis is somewhat simpler than ours, due to the fact that the distribution induced by the random projection operator is analytically more challenging that the subsampling operator.
Interestingly, they also show that one \emph{cannot} learn the subspace when observing only one coordinate per column, which may be related to why we require two compressive measurements, although of course the sampling paradigm is quite different. 

It may seem curious that the target rank $k$ is absent from the bound, but notice that there is an interaction between $k$, the bound on the column norms $\mu$, and the eigengap.
The target rank indirectly has some influence on the sample complexity $n$ via both the eigengap $\gamma_k$ and the length bound $\mu$. 

\ifthenelse{\equal{\version}{arxiv}}{
\begin{figure}[t]
\begin{center}
\includegraphics[scale=0.5]{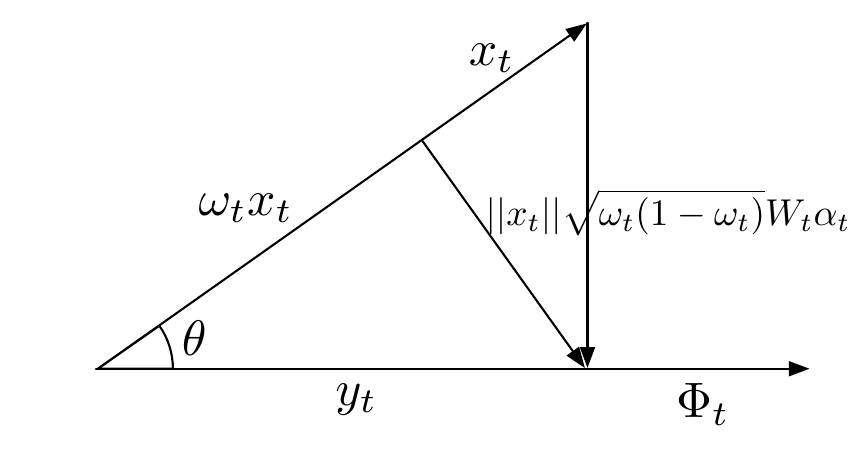}
\includegraphics[scale=0.3]{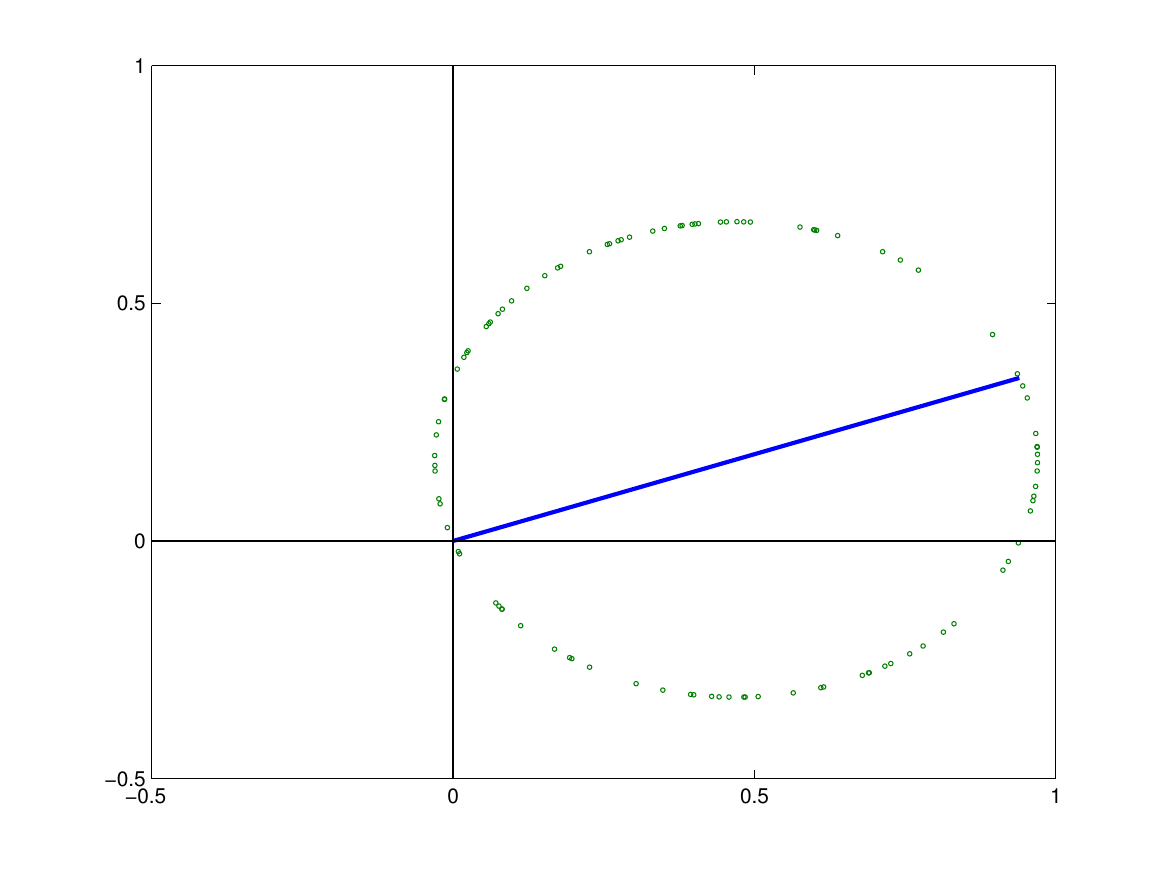}
\end{center}
\caption{Left: Geometry of the distribution of $y_t$. 
  Right: Distribution of $y_t$ (green dots) given $x_t$ (blue line).}
\label{fig:proj_geometry}
\end{figure}
}{
\begin{figure}[t]
\begin{center}
\includegraphics[scale=0.5]{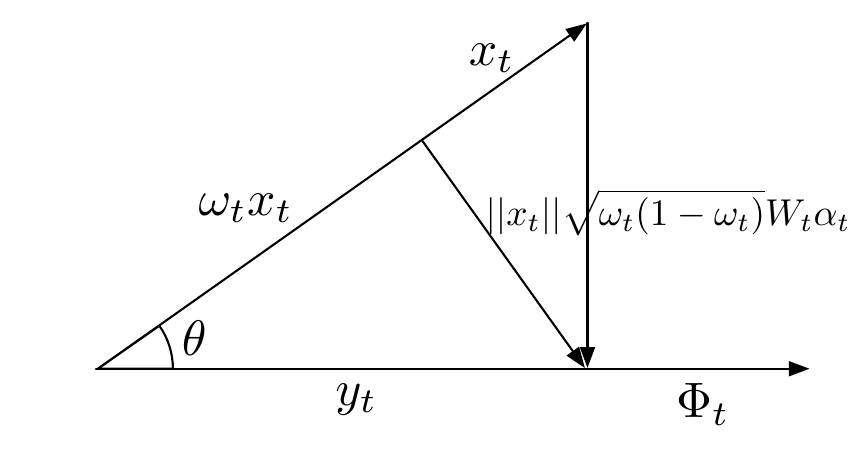}
\caption{Geometry of the distribution of $y_t$.}
\label{fig:geometry}
\end{center}
\vspace{-0.5cm}
\end{figure}

\begin{figure}[t]
\vspace{-0.25cm}
\begin{center}
\includegraphics[scale=0.3]{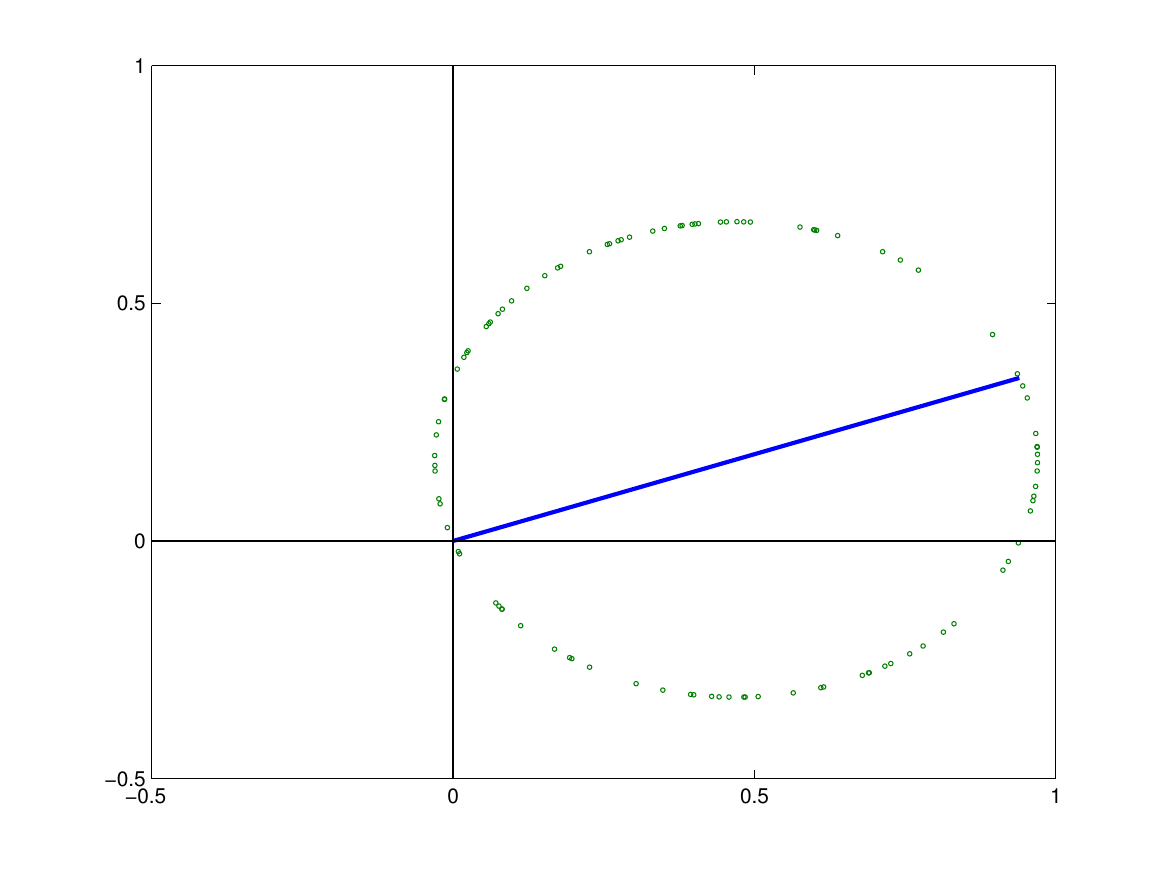}
\end{center}
\caption{Distribution of $y_t$ (green dots) given $x_t$ (blue line).}
\label{fig:proj_dist}
\vspace{-0.5cm}
\end{figure}
}

\begin{proof}
The main component of the proof is an exact characterization of the distribution of the observations $y_t, z_t$. 
Equipped with this distribution, the result is an application of two fairly well known results in the matrix perturbation literature. 
The first is the Matrix-Bernstein inequality, which analogously to the Bernstein inequality, formalizes the concentration of measure phenomenon. 
The second is the Davis-Kahan sine theorem, which characterizes the deviation of eigenvectors under a random perturbation of the matrix. 
We state both theorems here without proof:
\begin{theorem}[Matrix Bernstein Inequality~\cite{tropp2011user}]
Let $X_1, \ldots X_n$ be a sequence of independent, random, self-adjoint matrices with dimension $d$.
Assume that each random matrix satisfies:
\begin{align*}
\EE X_k = 0, \qquad ||X_k||_2 \le R \ \textrm{ a.s.}
\end{align*}
Then, for all $t \ge 0$,
\begin{align}
\PP\left(||\sum_k X_k||_2 \ge t\right) \le d \exp\left\{\frac{-t^2/2}{\sigma^2 + Rt/3}\right\}
\end{align}
where $\sigma^2 = ||\sum_{k} \EE (X_k^2)||_2$. 
\label{thm:matrix_hoeffding}
\end{theorem}
\begin{theorem}[Davis-Kahan Theorem~\cite{golub1996matrix,davis1970rotation}]
Let $A$ and $M$ be $d \times d$ self-adjoint matrices and denote the eigenvalues of $A$ as $\lambda_1 \ge \ldots \ge \lambda_d$.
Define $\Pi_k(A)$ (resp. $\Pi_k(M)$) to be the projection onto the top-$k$ eigenvectors of $A$.
Then:
\begin{align}
||\Pi_k(A) - \Pi_k(M)||_2 \le \frac{||A - M||_2}{\gamma_k(A)}
\end{align}
Where $\gamma_k(A) = \min_{j > k} |\lambda_k - \lambda_j|$ is the eigengap for $A$. 
\label{thm:davis_kahan}
\end{theorem}

Before diving into details of our proof, let us first capture some intuition.
Informally, the Davis-Kahan theorem says that the principal subspace of $\hat{\Sigma}$ is close to the principal subspace of $\Sigma$ provided that $||\hat{\Sigma} - \Sigma||_2$, once $\hat{\Sigma}$ is appropriately normalized, is small. 
At the same time, the Matrix Bernstein inequality shows that if we can write $\hat{\Sigma} - \Sigma$ as a sum of centered random matrices, then we can control the spectral norm deviation. 
Since $\Sigma = \frac{1}{n}\sum_{t=1}^n x_tx_t^T$ and by defining $C_t = x_tx_t^T$, it suffices to show that $\hat{C}_t$, defined in Equation~\ref{eq:chat_t}, is close to $C_t$. 
To see why this is true, we examine the vectors $y_t$ and $z_t$. 

\begin{figure*}
\begin{center}
\includegraphics[scale=\figsizeb]{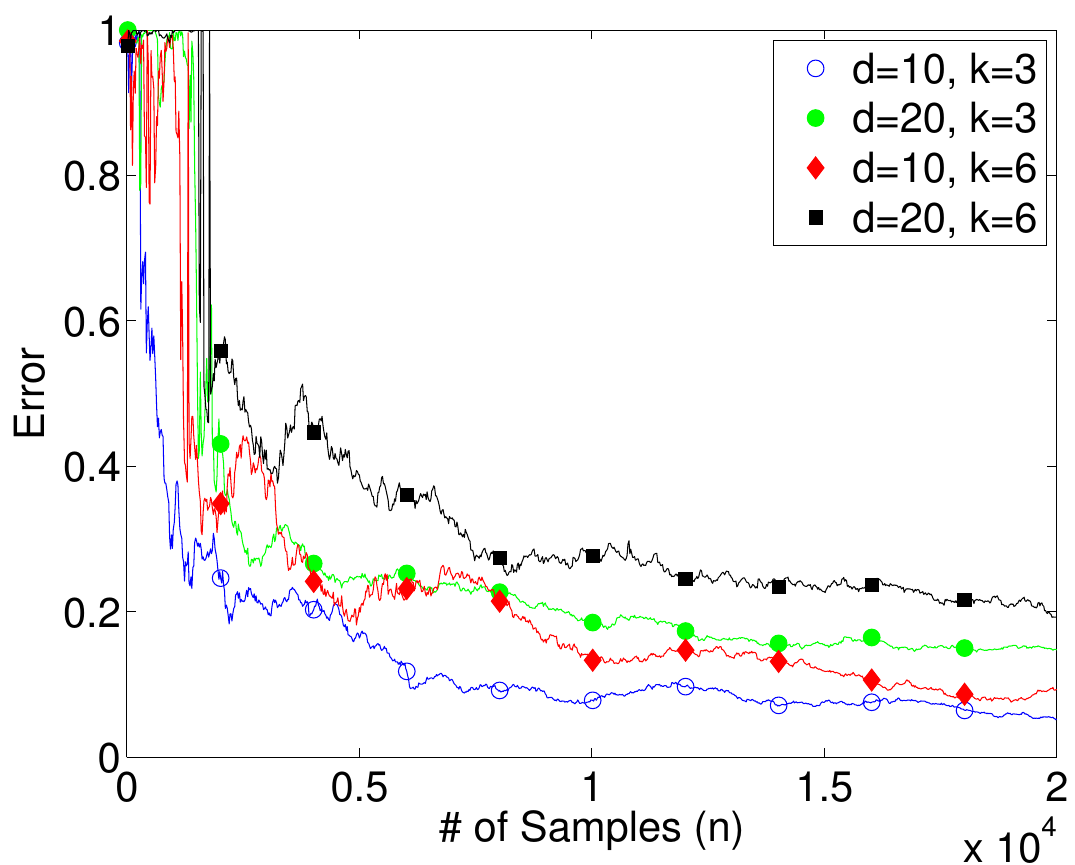}
\includegraphics[scale=\figsizeb]{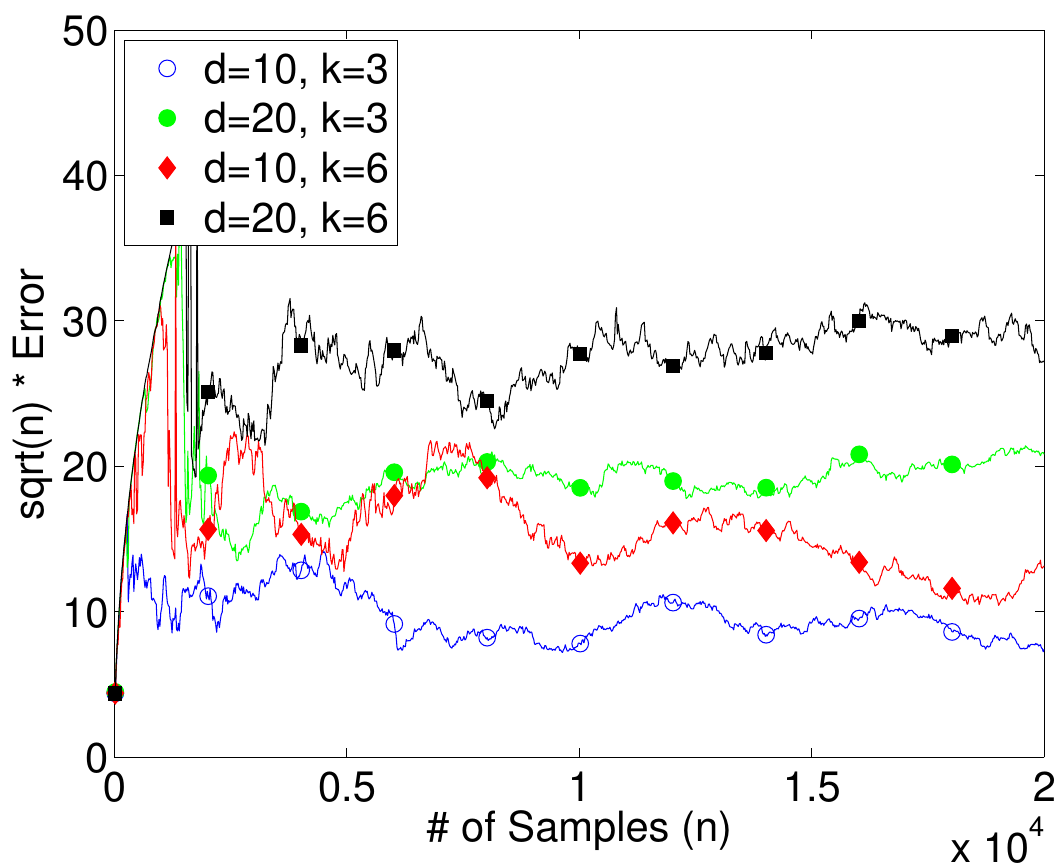}
\includegraphics[scale=\figsizeb]{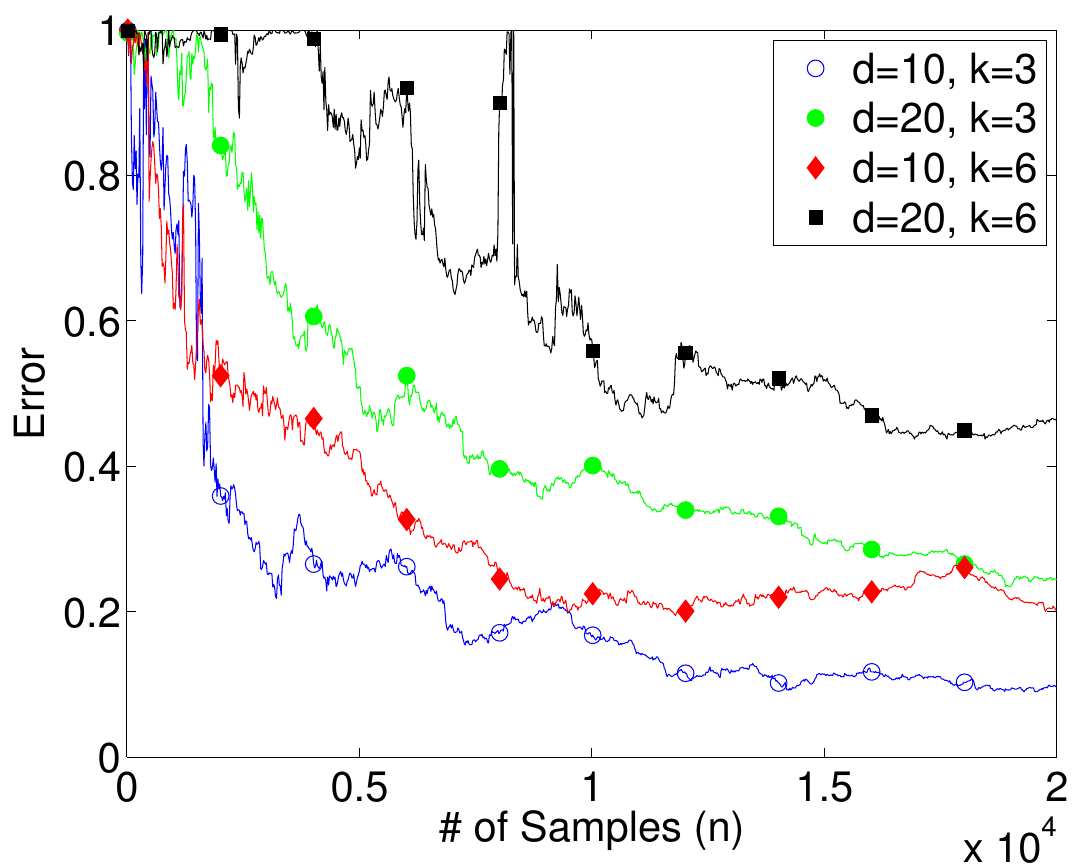}
\includegraphics[scale=\figsizeb]{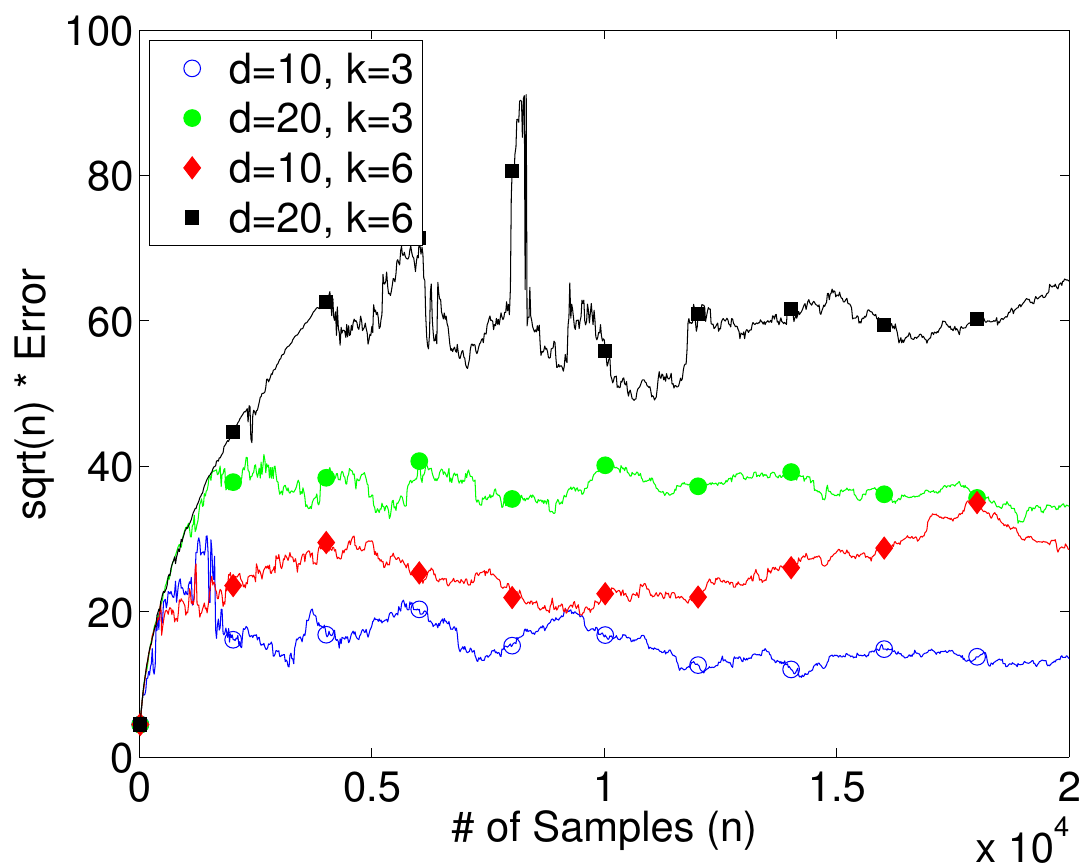}\\
\includegraphics[scale=\figsizeb]{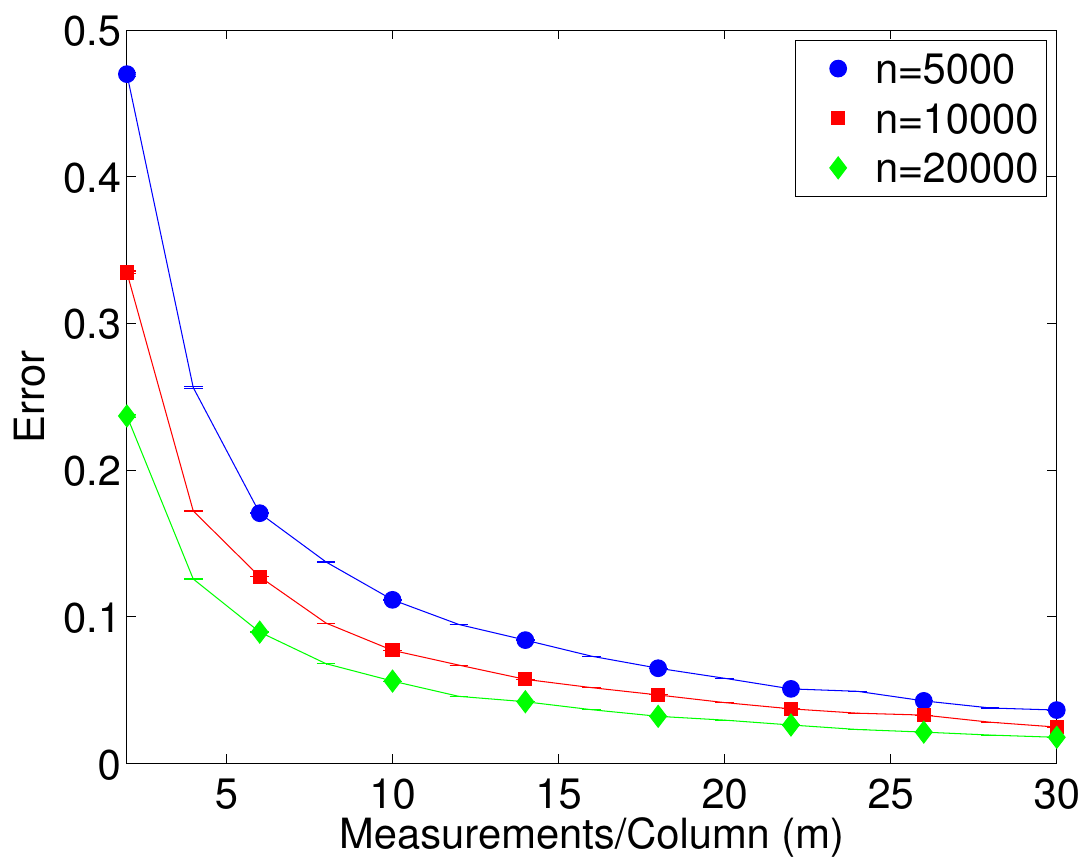}
\includegraphics[scale=\figsizeb]{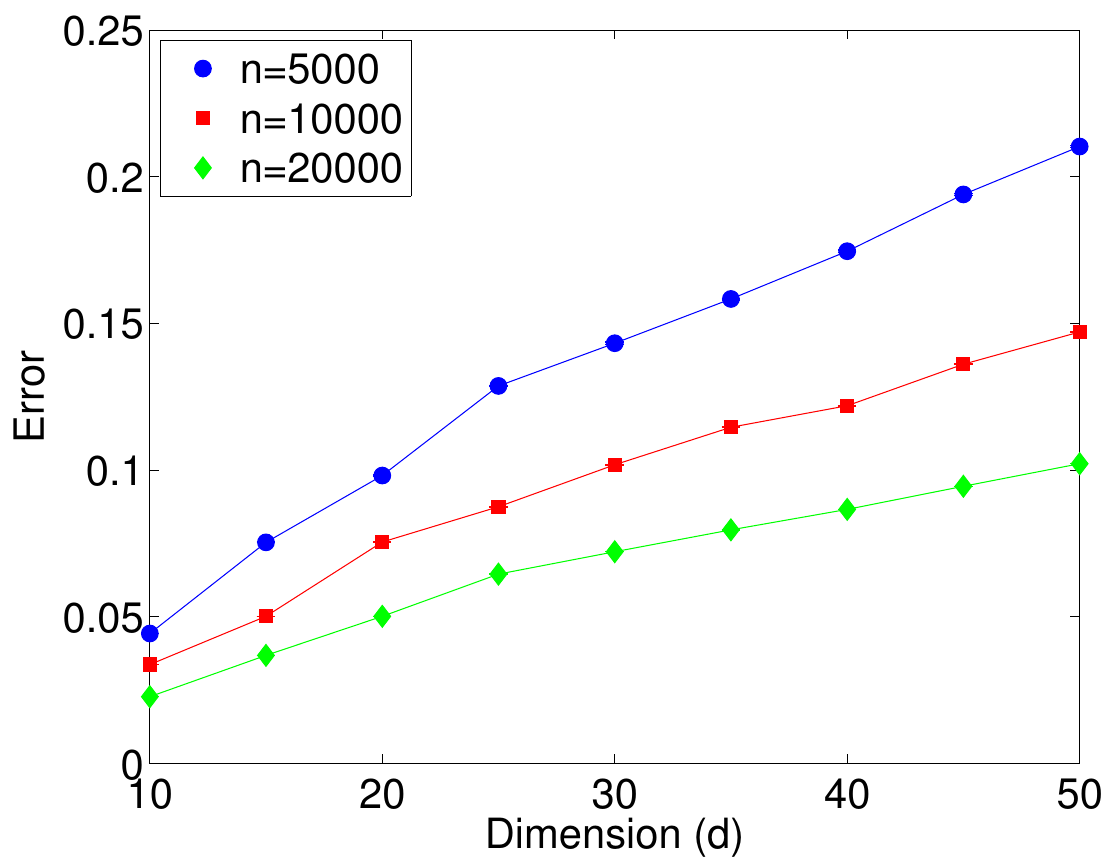}
\includegraphics[scale=\figsizeb]{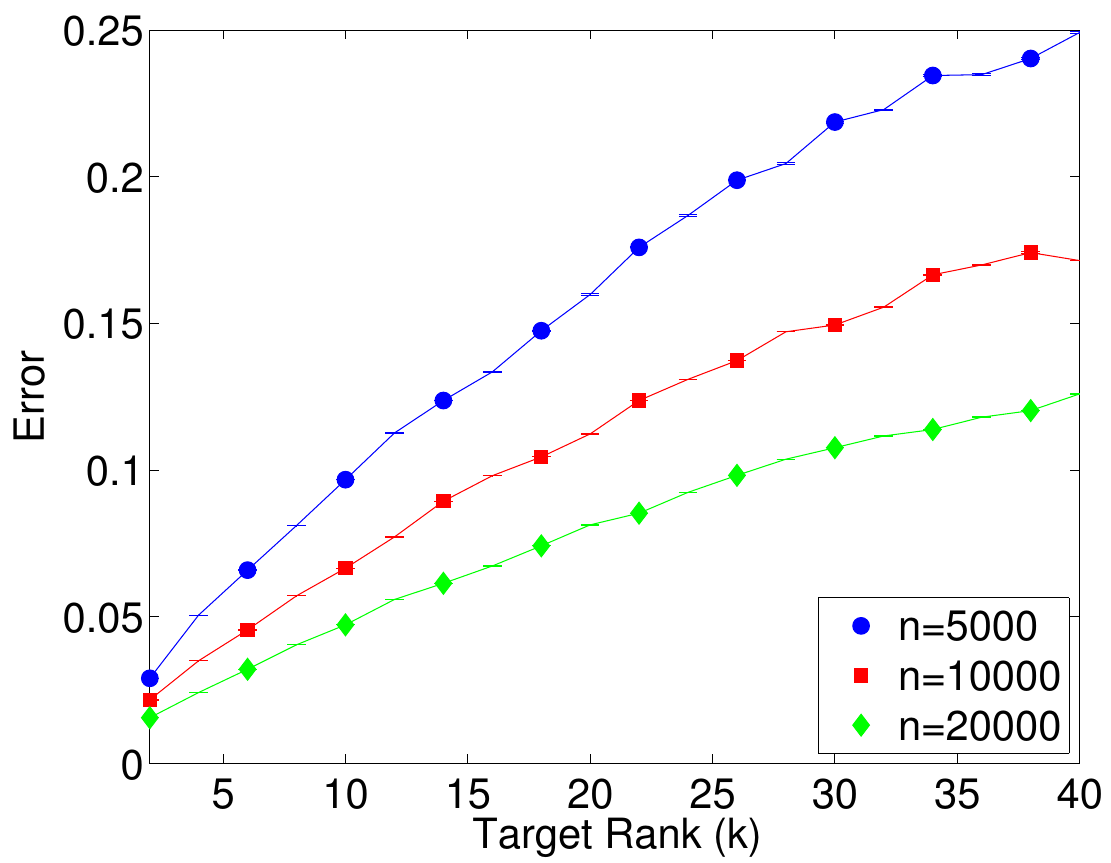}
\includegraphics[scale=\figsizeb]{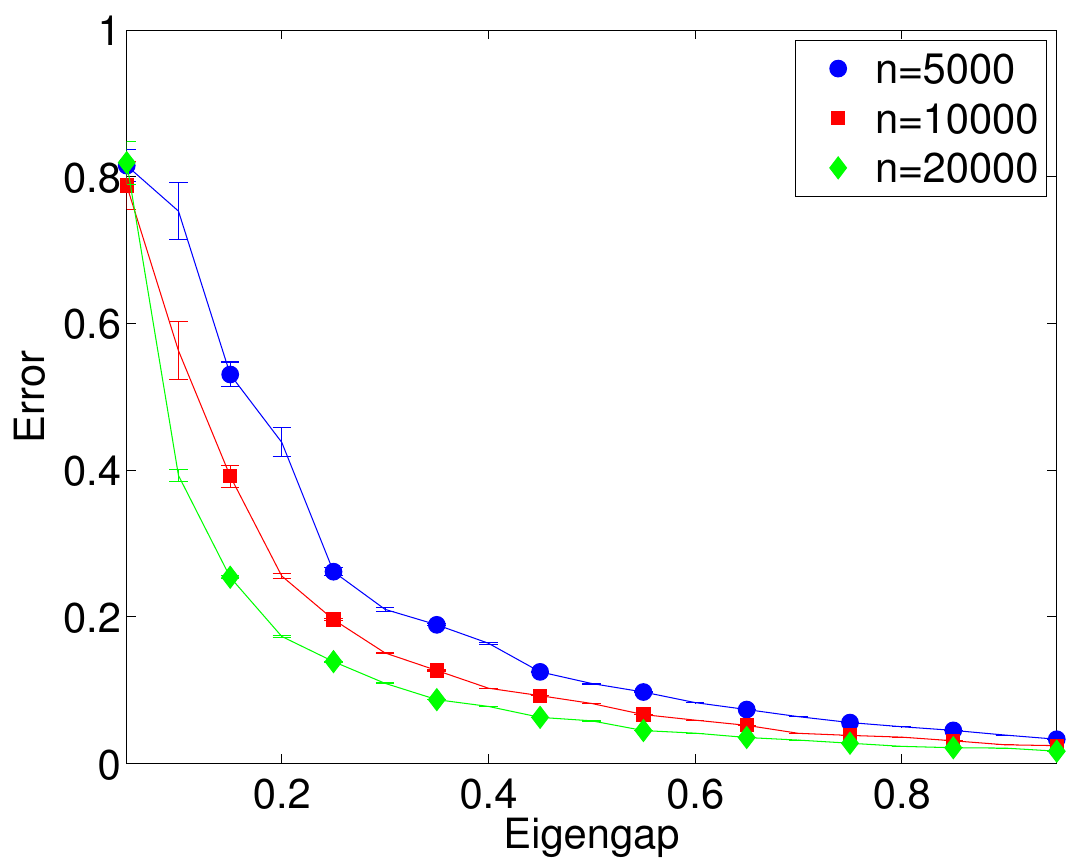}
\end{center}
\caption{Top row: Rates of convergence alongside rescaled rates  $\sqrt{n}\times ||\hat{\Pi} - \Pi||_2$ across a number of problem settings.
Left two plots show exactly low rank case while the right two plots show low rank approximation of high rank matrices.
Bottom row: Influence of other problem parameters on error.
From left to right: number of measurements per column ($m$), problem dimension ($d$), target rank $(k)$, and eigengap ($\gamma_k$).}
\label{fig:convergence}
\vspace{-0.5cm}
\end{figure*}

Let us focus on $y_t$. 
An equivalent way to analyze $y_t$ is to consider $\Phi_t$ fixed and draw the direction of $x_t$ uniformly. 
By rotational invariance, we can think of $\Phi_t$ as projecting onto the first $m$ standard basis elements, in which case it is easy to see that that $||y_t||_2^2$ has the same distribution as $\omega_t ||x_t||_2^2$ where $\omega_t \sim \textrm{Beta}(\frac{m}{2}, \frac{d-m}{2})$\footnote{$\omega_t$ can also be written as the ratio $\frac{a}{a+b}$ where $a \sim \chi_m^2, b \sim \chi^2_{d-m}$.}.

To capture the distribution of $y_t$, we need a more geometric argument
\ifthenelse{\equal{\version}{arxiv}}{(See the left panel of Figure~\ref{fig:proj_geometry})}{(See Figure~\ref{fig:geometry})}. 
The angle $\theta$ between $x_t$ and $\Phi_t$ is $\textrm{cos}^{-1}(\sqrt{\omega_t})$, which means that the magnitude of $y_t$ in the direction of $x_t$ is $||y_t||_2\cos(\theta) = \omega_t ||x_t||_2$.
Moreover, the magnitude of $y_t$ in the orthogonal direction is $||x_t||_2 \sqrt{\omega_t(1-\omega_t)}$ and this direction is chosen uniformly at random (subject to being orthogonal to $x_t$). 
Thus $y_t$ is distributed as:
\[
\omega_t x_t + ||x_t||_2\sqrt{\omega_t(1-\omega_t)} W_t \alpha_t
\]
where $W_t \in \RR^{d \times d-1}$ is a basis for the set of vectors orthogonal to $x_t$ and $\alpha_t$ is drawn uniformly from the unit sphere in $\RR^{d-1}$ and independently from $\omega_t$. 

Analogously $z_t$ is distributed as:
\[
\eta_t x_t + ||x_t||_2 \sqrt{\eta_t(1-\eta_t)} W_t \beta_t
\]
Where the random variables $\eta_t, \beta_t$ are independent from each other and from $\omega_t, \alpha_t$. 

Since $\alpha_t, \beta_t$ are drawn uniformly on the sphere, it is clear that $\EE\alpha_t = \EE\beta_t = 0$, so that $\EE y_t = \EE z_t = \frac{m}{d} x_t$. 
Thus $\frac{d^2}{m^2} \hat{C}_t$ is an unbiased estimator for $C_t$.
We are interested in approximating $\Sigma = \frac{1}{n}\sum_{t} C_t$, so the appropriate random variables to control are $X_t = \frac{d^2}{nm^2} \hat{C}_t - \frac{1}{n} C_t$. 

To apply Theorem~\ref{thm:matrix_hoeffding}, we just need to compute the variance $\sigma^2$ and verify that $||X_t||_2$ is bounded. 
The latter is immediate; since $\Phi_t, \Psi_t$ are projection operators, it must be the case that $||y_t||_2,||z_t||_2 \le ||x_t||_2$, so that we can set $R = 2 \frac{d^2}{nm^2} \mu$ (assuming $m \le d$).

\ifthenelse{\equal{\version}{arxiv}}{
Some careful calculations, detailed in the appendix, show that the variance can be bounded by:
}{
Some careful calculations, detailed in the supplementary material~\cite{krishnamurthy2014subspace}, show that the variance can be bounded by:
}
\[
\sigma^2 \le 7 \frac{\mu^2d }{mn}.
\]
We can now apply the inequality, which, once inverted, says that with probability $\ge 1-\delta$:
\begin{align}
||\frac{d^2}{n m^2}\hat{\Sigma} - \Sigma||_2 \le \sqrt{\frac{14 \mu^2 d}{nm}\log(d/\delta)} + \frac{2}{3}\frac{\mu d^2}{nm^2}\log(d/\delta).
\end{align}
Plugging this into the bound in Theorem~\ref{thm:davis_kahan} establishes Equation~\ref{eq:err_bound}.
Equation~\ref{eq:sample_complexity} is just a rearrangement of Equation~\ref{eq:err_bound}.
\end{proof}

%% file: experiments.tex
\section{Experiments}
\label{sec:experiments}

We complement our theoretical study of the compressive subspace learning algorithm with a number of simulations. 
Recall that our main goal is to capture the dependence between the error $\epsilon$ and the number of columns $n$.
We do not believe that our results are tight in their dependence on the other parameters $d, m, \gamma_k, \mu$. 
However we do verify qualitatively the influence of the other parameters on the error. 

Our first simulation is a comparison between CSL and the compressive singular value decomposition algorithm of Halko \emph{et. al}~\cite{halko2011finding}.
Their algorithm simply projects the columns of $X$ onto a fixed $m$-dimensional subspace of $\RR^d$ and uses principal subspace of $Y = \Phi X$ as the estimate for $\Pi$. 
The main difference between their algorithm and ours is that we use a different $m$-dimensional projection per column, which leads to significant improvement in performance as demonstrated in Figure~\ref{fig:hmt_compare}.
In that figure, we plot the error $||\hat{\Pi} - \Pi||_2$ as a function of the number of columns $n$, for two different choices of $m$.
The matrix rank and the target rank $k$ is $2$ and $d=20$. 

As predicted by Theorem~\ref{thm:main}, the error for our algorithm converges quickly to zero with $n$, with better convergence for larger $m$.
On the other hand, the HMT algorithm does not enjoy any performance improvements as $n$ increases, which is also predicted by their theoretical results. 
Thus we empirically see the averaging effect that is formalized by our theory. 

We also verify that the \emph{rate} of convergence in Theorem~\ref{thm:main} is substantiated by empirical simulations. 
To this end, in Figure~\ref{fig:convergence}, we plot both the error alongside the rescaled error $\sqrt{n} ||\hat{\Pi} - \Pi||_2$ as a function of $n$ across several problem settings. 
In the left two plots we consider the exactly low rank case, while on the right two plots we consider matrices that have some non-zero eigengap $\gamma_k$, but are not exactly low rank. 

The first thing to notice is that in all simulations, $\sqrt{n} ||\hat{\Pi} - \Pi||_2$ does appear to level out to some constant value, which demonstrates that our algorithm does converge at $n^{-1/2}$ rate. 
Moreover, in comparing between the trials, we see that increasing the dimensionality degrades the performance of the algorithm.
On the other hand, increasing the target rank $k$ seems to have less influence. 

We study the effect of $d, m, k$ and $\gamma_k$ in the bottom row of Figure~\ref{fig:convergence}.
Our interpretation of the results is somewhat more qualitative, as we do not expect our theory to precisely capture the dependence between the error and these parameters. 
In the first figure, we see that increasing the number of measurements per column $m$ significantly improves the performance of the algorithm, as one might expect. 
It does appear that the dependence is $\epsilon \asymp 1/m$, rather than the inverse-quadratic dependence in Theorem~\ref{thm:main}.
In terms of problem dimension $d$, plotted in the second figure, the dependence appears to be linear rather than quadratic as predicted by Theorem~\ref{thm:main}.

As we mentioned earlier, there is no explicit dependence on the target rank $k$ in Theorem~\ref{thm:main}, although it does play some role indirectly through $\gamma_k$ and $\mu$. 
However, in the third plot from the left of the bottom row of Figure~\ref{fig:convergence}, we clearly see a linear dependence between $k$ and $\epsilon$.
We suspect that a more careful analysis of our algorithm can explicitly introduce the dependence on $k$ while possibly removing one factor of $d$, leading to a bound that is more reminiscent of existing matrix completion and compression results~\cite{halko2011finding,recht2011simpler,negahban2012restricted}.

Lastly we plot the effect of the eigengap $\gamma_k$ on the error.
We comment on two main effects: (1) increasing $\gamma_k$ improves the performance of the algorithm, and (2) for fixed eigengap, the performance improves with $n$. 
This qualitatively justifies $\gamma_k$ in the denominator of Equation~\ref{eq:err_bound}.

%% file: conclusion.tex
\section{Conclusion}
\label{sec:conclusion}

In this manuscript, we demonstrate how one can approximate the principal subspace of a data matrix from very few compressive measurements per column.
The main insight is that by using an independent random compression operator on each column, we can effectively take averages across columns, which preserves the signal but diminishes the noise stemming from the compression. 

These results are practically and theoretically appealing yet several challenges still remain.
First, our simulations suggest that the error should scale linearly with $d$ and $k$ and inversely linearly with $m$, yet our error bound has worse dependences.
We would like to improve these dependences so that the theoretical results are more predictive of our experimental findings.
Secondly, as we mentioned before, in theory, our algorithm requires at least two compressive measurements per column, although we did not observe such a requirement in simulation.
While Gonen \emph{et. al} proved that such a requirement is necessary in the missing data setting, their justification does not immediately carry over to our setting~\cite{gonen2014sample}.
It would be worthwhile to understand the differences between these sampling paradigms and to show either that two compressive measurements per column are necessary or that one suffices.

Lastly, it is not clear what the fundamental limits are for the compressive subspace learning problem.
Given the compressed measurements $y_t, z_t$ for each column $x_t$, is there a lower bound on the error achievable by \emph{any} algorithm?
Is there an algorithm that achieves, or nearly achieves, this bound?

We hope to address these questions in future work.

%% file: appendix.tex
\section{Bounding the Variance}
Here we bound the variance:
\begin{align*}
\sigma^2 = \left\| \sum_{t=1}^n\EE X_t^2\right\|, \qquad X_t = \frac{d^2}{2m^2n}\left(y_t z_t^T + z_ty_t^T\right) - \frac{1}{n} x_tx_t^T.
\end{align*}
Recall that:
\begin{align*}
y_t &= \omega_t x_t + \|x_t\|\sqrt{\omega_t(1-\omega_t)}W_t\alpha_t\\
z_t &= \eta_t x_t + \|x_t\|\sqrt{\eta_t(1-\eta_t)}W_t \beta_t
\end{align*}
where $\omega_t,\eta_t \sim \textrm{Beta}(\frac{m}{2}, \frac{d-m}{2})$, $W_t \in \RR^{d \times d-1}$ is a basis for the set of vectors orthogonal to $x_t$ and $\alpha_t, \beta_t$ are drawn uniformly from the unit sphere in $\RR^{d-1}$. 
All random variables are independent. 

Notice that $\EE\omega_t^2 = \EE \eta_t^2 = \frac{m(m+2)}{d(d+2)}$, $\EE\alpha=\EE\beta=0$, and $\EE \alpha_t\alpha_t^T\beta_t\beta_t^T = \frac{I_{d-1}}{(d-1)^2}$, so that:
\begin{align*}
\EE y_tz_t^Ty_tz_t^T &= \EE\left( \omega_t \eta_t \|x_t^2\| + \|x_t\|^2 \sqrt{\omega_t(1-\omega_t)}\sqrt{\eta_t(1-\eta_t)} \alpha_t^T\beta_t\right) \times \\
& \times \left(\omega_t \eta_t x_tx_t^T + \omega_t \|x_t\|\sqrt{\eta_t(1-\eta_t)} x_t\beta_t^TW_t^T + \eta_t \|x_t\|\sqrt{\omega_t(1-\omega_t)} W_t\alpha_T x_t^T\right.\\
& \left.+ \|x_t\|^2\sqrt{\omega_t(1-\omega_t)}\sqrt{\eta_t(1-\eta_t)} W_t\alpha_t\beta_t^TW_t^T\right)\\
& = \|x_t\|^2\left(\frac{m(m+2)}{d(d+2)}\right)^2 x_tx_t^T + \|x_t\|^4\left( \frac{m(d-m)}{d(d+2)(d-1)}\right)^2 W_tW_t^T.
\end{align*}
This follows since the last two terms in the second line make no contribution to the sum.

We also have:
\begin{align*}
& \EE y_tz_t^T z_t y_t^T \\
&= \EE \left(\eta_t^2\|x_t\|^2 + \|x_t\|^2(\eta_t-\eta_t^2)\|\beta_t\|^2 \right) \times \\
&\times \left(\omega_t^2 x_tx_t^T + \omega_t \|x_t\| \sqrt{\omega_t(1-\omega_t)} (x_t \alpha_t^T W_t^T + W_t \alpha_t x_t^T )
+ \|x_t\|^2 (\omega_t-\omega_t^2)W_t\alpha_t\alpha_t^T W_t^T\right) \\
& = \EE \left(\eta_t\|x_t\|^2 \right) \times \left(\omega_t^2 x_tx_t^T + \|x_t\|^2 (\omega_t-\omega_t^2)W_t\alpha_t\alpha_t^T W_t^T\right) \\
& = \EE\left( \omega_t^2\eta_t \|x_t\|^2 x_tx_t^T + \eta_t(\omega_t - \omega_t^2)\|x_t\|^4W_t\alpha_t\alpha_t^TW_t^T\right)\\
& = \|x_t\|^2 \frac{m^2(m+2)}{d^2(d+2)}x_tx_t^T + \|x_t\|^4\frac{m^2(d-m)}{d^2(d+2)(d-1)}W_tW_t^T.
\end{align*}
The first equality expands definitions, while in the second we use
that since $\beta_t$ is distributed uniformly on the unit sphere, we
have $\|\beta_t\|^2 = 1$ with probability $1$.  The second equality
also uses the fact that the terms that are linear in $\alpha_t$ are
zero in expectation. The rest of the calculation follows by
independence and by the definitions of the random variables.

Thus we can bound the variance by:
\begin{align*}
\sigma^2 &= \frac{d^4}{m^4n^2} \left\| \sum_{t=1}^n \frac{1}{2} \EE y_tz_t^Ty_tz_t^T + \frac{1}{2}\EE y_tz_t^Tz_ty_t^T- \frac{m^4}{d^4} x_tx_t^Tx_tx_t^T\right\|\\
& \le \frac{d^4}{m^4n^2} \sum_{t=1}^n \left\| \frac{1}{2} \EE y_tz_t^Ty_tz_t^T + \frac{1}{2}\EE y_tz_t^Tz_ty_t^T- \frac{m^4}{d^4} x_tx_t^Tx_tx_t^T\right\|.
\end{align*}
Expanding the expectations, there are three terms involving $x_tx_t^T$ and two terms involving $W_tW_t^T$. These terms are,
\begin{align*}
T_{1,t} = \|x_t\|^2 \left(\frac{1}{2}\left(\frac{m(m+2)}{d(d+2)}\right)^2 + \frac{1}{2}\frac{m^2(m+2)}{d^2(d+2)} - \frac{m^4}{d^4}\right)\|x_tx_t^T\|\\
T_{2,t}= \|x_t\|^4 \left(\frac{1}{2}\left(\frac{m(d-m)}{d(d+2)(d-1)}\right)^2 + \frac{1}{2}\frac{m^2(d-m)}{d^2(d+2)(d-1)}\right)\|W_tW_t^T\|.
\end{align*}
By the triangle inequality
\begin{align*}
  \sigma^2 \le \frac{d^4}{n^2 m^2}\sum_{t=1}^n T_{1,t} + T_{2,t}.
\end{align*}
We now proceed to bound the terms $T_{1,t}$ and $T_{2,t}$. 
The main point is that this expression is actually $O(\frac{\mu^2 d}{mn})$, which gives us the variance bound used in our theorem.
Clearly we have $\|x_tx_t^T\| = \|x_t\|^2 \le \mu$ and $\|W_tW_t^T\| = 1$, so to conclude we need to control the rational functions. 
First,
\begin{align*}
T_{1,t} &\le \mu^2 \left(\frac{1}{2}\left(\frac{m(m+2)}{d(d+2)}\right)^2 + \frac{1}{2}\frac{m^2(m+2)}{d^2(d+2)} - \frac{m^4}{d^4}\right)\\
& = \frac{\mu^2 m^2}{d^2}\left(\frac{(m+2)^2}{2(d+2)^2} + \frac{m+2}{2(d+2)} - \frac{m^2}{d^2}\right)\\
& = \frac{\mu^2m^2}{d^2}\left( \frac{d^2(m+2)^2 - m^2(d+2)^2}{2d^2(d+2)^2} + \frac{d^2(m+2) - m^2(d+2)}{2d^2(d+2)}\right)\\
& \le \frac{\mu^2m^2}{d^2}\left( \frac{4d^2m + 4d^2}{2d^2(d+2)^2} + \frac{d^2m + 2d^2}{2d^2(d+2)}\right)\\
& \le \frac{4\mu^2m^3}{d^4} + \frac{3\mu^2m^3}{2d^3}.
\end{align*}

We also have,
\begin{align*}
T_{2,t} &\le \frac{\mu^2}{2}\frac{m^2(d-m)}{d^2(d+2)(d-1)}\left(\frac{d-m}{(d+2)(d-1)} + 1\right)\\
& \le \mu^2 \frac{m^2(d-m)}{d^2(d+2)(d-1)}  \le\mu^2 \frac{m^2}{d^3}.
\end{align*}
Putting things together, leads to an upper bound on the variance term $\sigma^2$,
\begin{align*}
\sigma^2 &\le \frac{\mu^2d^4}{n m^4} \left(\frac{4 m^3}{d^4} + \frac{3m^3}{2d^3} + \frac{m^2}{d^3}\right)
\le \frac{7\mu^2 d}{nm}.
\end{align*}


